\documentclass{article}
\usepackage{ijcai17}

\usepackage{times}

\usepackage{times,helvet,courier}
\usepackage{graphicx,color,balance}
\usepackage{amssymb,amsmath,array,amsfonts,mathrsfs}
\usepackage[ruled, vlined]{algorithm2e}
\usepackage{enumitem}
\usepackage{pgf}
\usepackage{multicol}
\usepackage{fixltx2e}
\usepackage{placeins}
\usepackage{tikz}
\usepackage{url}
\usepackage{tikzscale}
\usetikzlibrary{arrows} 
\usetikzlibrary{shapes.geometric}
\usetikzlibrary{shapes.arrows}

\usepackage[font=small,skip=0pt]{caption}
\usepackage{xcolor}
\usepackage{newverbs}

\newverbcommand{\cverb}{\color{red}}{}
\newverbcommand{\bverb} {\begin{lrbox}{\verbbox}}{\end{lrbox}\colorbox{gray!30}{\box\verbbox}}

\SetKwFor{Loop}{Loop}{}{EndLoop}

\usetikzlibrary{arrows,automata}
\usetikzlibrary{fit}					
\usetikzlibrary{backgrounds}	
\usetikzlibrary{decorations.pathmorphing,backgrounds,positioning,fit,petri}

\usepackage{ifthen}

\usepackage[textsize=scriptsize,textwidth=1cm]{todonotes}

\usepackage{braket}

\begin{document}

\title{Towards Providing Explanations for AI Planner Decisions}
\author{Rita Borgo, Michael Cashmore, Daniele Magazzeni\\
King's College London\\
\textit{firstname.lastname@kcl.ac.uk}
}
\maketitle

\begin{abstract}
In order to engender trust in AI, humans must understand what an AI system is
trying to achieve, and why. To overcome this problem, the underlying AI process must produce justifications and explanations that are both transparent and comprehensible to the user.
AI Planning is well placed to be able to address this challenge.

In this paper we present a methodology to provide initial explanations for the decisions made by the planner. Explanations are created by allowing the user to suggest alternative actions in plans and then compare the resulting plans with the one found by the planner.
The methodology is implemented in the new \textsc{XAI-Plan} framework.

\end{abstract}

\section{Introduction}
Artificial Intelligence technologies are increasingly ubiquitous in modern society and have the potential to fundamentally change all aspects of our lives. At the same time, increasing concerns are being raised about the transparency and accountability of AI systems. 

The European Parliament resolved in 2017 to require AI systems to follow a principle of transparency, meaning they should be able to justify their decisions in a manner comprehensible
to a human.
Similarly, legal measures are being adopted to provide individuals affected by automated decision-making with a "right to explanation", as referred to in the recent EU General Data Protection Regulation (GDPR), in place from May 2018.

In order to engender trust in AI, humans must understand what an AI system is trying to achieve, and why. To overcome this problem, the underlying AI process must produce justifications and explanations that are both transparent and comprehensible to the user. 

In 2016 DARPA launched the Explainable AI program, and since then the AI research community has been looking at this challenge with growing interest\footnote{See Workshops on Explainable AI at IJCAI-17 and IJCAI-18.}, mainly trying to explain or verify neural networks.
While much current attention is focussed on the recent advancements in data-driven AI (e.g., Machine Learning, Deep Learning), model-based AI such as Planning is well placed to be able to address the challenges of transparency and explainability.

In AI Planning, most previous work within the realm of explainability aims to help humans understand the elements of a plan suggested by the system (e.g. \cite{Sohrabi}, ). This involves the transformation of planner output into forms that non-expert users can understand and the description of causal and temporal relations between plan steps. However, describing how a single plan works is a different problem than explaining why the planner suggests a particular plan. This requires new fundamental algorithms for generating explanations. Now, planning is being used in new critical domains (e.g., smart grid \cite{smartgridSylvie,UCP}, urban/air traffic control \cite{vallatiUTC,morrisPlane}, mining \cite{pearceMining}, underwater robotics \cite{auv}) and handles temporal constraints, numeric resources and continuous change \cite{dino16}, resulting in more complex plans. In this regard, explanation-generating algorithms will be instrumental in developing more robust systems for use in these critical domains.

In \cite{xaip} a roadmap for addressing \textit{Explainable Planning} is proposed, which identifies a list of questions that planning systems should be able to answer, both offline before the plan is approved as well as online during plan execution. In this paper we tackle some of these questions, and in particular we focus on what we think are the most common questions a user would ask when confronted with a plan: ``\textit{Why would you do this rather than that?}", and ``\textit{Why is what you suggest more efficient than something else?}".
Confronted with a question or an alternative indicated by a human user, the explanation should be a demonstration that the alternative would prevent the generation of a valid plan, or at least be no better than the existing plan. This would be a justification for the choices made by the planner. An important side effect of such an approach is that this interaction between the user and the planner enhances mixed-initiative planning, and it might be the case that the suggestion made by the user actually improves the final plan. When times and real numbers are taken into account, as it is the case in many real-world scenarios, one cannot expect optimal plans from a planner, and hence the knowledge of the domain expert should be considered in the planning process.

In this paper we present a methodology to provide explanation for the decisions made by the planner. Explanations are created by allowing the user to explore alternative actions in the plans and then compare the resulting plans with the one found by the planner, using different metrics.
We implemented the methodology in the new \textsc{XAI-Plan} framework. The methodology is domain-independent and is agnostic about the planning system used. We evaluated the framework in a number of domains using ROSPlan \cite{rosplan15}.

The paper is structured as follows. In Section \ref{sec:relwork} we provide a brief overview of related work. In Section \ref{sec:questions} we present the methodology and in Section \ref{sec:examples} we demonstrate the methodology with a working example. In Section \ref{sec:implementation} we describe the implementation of the {\sc XAI-Plan} framework, and Section \ref{sec:conclusion} concludes the paper.

\section{Related Work}\label{sec:relwork}

\textit{Plan Explanation} is an area of planning where the main goal is to help humans understand the plans produced by the planners (e.g., \cite{Sohrabi,biundo,biundospoken}). This involves the translation of plans to forms that humans can easily understand and the design of interfaces that help this understanding.
Relevant works in robotics include \cite{shah,veloso1}. 
Similar work focuses on generating diverse solutions when user preferences are not known~\cite{ngu12}, or top-k solutions~\cite{kat18}. While providing the user with more choice, this does not necessarily provide explanation, nor prevent the user asking why a particular plan is selected.

Kambhampati and his team focus on the important scenario where humans and the agent have different models of the world. Explanations in that context must handle the issue of \textit{model reconciliation}~\cite{rao1,rao3}. In the same context, \textit{Plan Explicability}~\cite{rao2} focuses on human interpretation of plans. 
In this stream of works, the focus is on optimal plans in classical planning, which might differ because of the different models used to generate them.  
We focus on more expressive domains where the model is well defined, but the resulting state space is too vast and complex. In cases where the model is sufficiently complex, it is not possible to provide explanations that can be well understood in the form of model reconciliation. In this line of research, relevant works include \cite{smith12,langley17,xaip}.

\section{Explanations for Planner Decisions}\label{sec:questions}

When confronted with a plan generated by a planner, a common question the user might ask is: "\textit{Why does the plan contain this action rather than this other action that I would expect?}". Indeed, it should be noted that the support of AI is even more relevant when the AI suggests to do something \textit{different} from what the user would do (and in particular a domain expert). At the same time such a different action plan would need an explanation before the user can be confident on its effectiveness and approve the plan.

For an effective explanation, rewriting the steps of the planning algorithm in natural language is not what is required. Nor is it very helpful to provide the heuristic evaluations of the states selected by the planner when searching for a plan~\cite{xaip}. 

Rather, we argue that what can justify the selection of a set of actions by a planner is that this set of actions proves to be better, or at least no worse, than the set of actions the user would select. To this aim, a framework for providing explanations should allow the user to explore alternative actions in the plan and then compare the resulting plans with the one found by the planner. Different metrics can be used to evaluate the quality of different plans. Such an approach would increase the confidence of the user in the planner and would give him/her evidence for accepting or rejecting plans.

In this work, we use this approach to tackle the first three questions considered in the roadmap for explainable planning proposed in \cite{xaip}:
\begin{itemize}
\item [Q1] Why did you do that?
\item [Q2] Why didn't you do something else (that I would have done)?
\item [Q3] Why is what you propose to do more efficient than something else (that I would have done)?
\end{itemize}

\begin{figure*}[thb]
  \centering
\begin{tikzpicture}[>=latex]
   
  \begin{scope}


  \filldraw[black] (0,0) circle (2pt) node[anchor=south] {s};
  \node (p0_d) at (-0.4,-0.4) {A};
  
  \filldraw[black] (-0.5,-1.0) circle (2pt) node[anchor=east] {$s_1$};
  
  \draw[<-] (-0.4,-0.9) -- (0.0,0.0);
  \draw[<-] (0.4,-0.9) -- (0.0,0.0);

  \filldraw[black] (0.5,-1.0) circle (2pt) node[anchor=west] {$r_1$};
\node (p0_d) at (0.4,-0.4) {B};
\draw [->,dashed,blue] (0.5,-1.0) to [bend left=45] (0.0,-0.2);
  
  \filldraw[black] (-0.5,-4.3) circle (2pt) node[anchor=north] {$g_{A}$};

\draw [->,decorate,decoration=snake] (-0.5,-1.0) -- (-0.5,-4.3);

  \filldraw[black] (3.5,0) circle (2pt) node[anchor=south] {s};
  \node (p0_d) at (3.1,-0.4) {A};
  
  \filldraw[black] (3,-1.0) circle (2pt) node[anchor=east] {$s_1$};
  
  \draw[<-] (3.1,-0.9) -- (3.5,0.0);
  \draw[<-] (3.9,-0.9) -- (3.5,0.0);

  \filldraw[black] (4.0,-1.0) circle (2pt) node[anchor=west] {$r_1$};
\node (p0_d) at (3.9,-0.4) {B};

  \filldraw[black] (2.75,-2.0) circle (2pt) node[anchor=east] {$s_2$};
  \draw[<-] (2.75,-1.9) -- (3.0,-1.0);
  \node (p0_d) at (2.65,-1.5) {$\alpha_1$};

  \filldraw[black] (4.25,-2.0) circle (2pt) node[anchor=west] {$r_2$};
  \draw[<-] (4.25,-1.9) -- (4.0,-1.0);
  \node (p0_d) at (4.35,-1.5) {$\beta_1$};
  
  \node (dotsl) at (2.55,-2.72) {$\cdots$};
  \node (dotsl) at (4.4,-2.72) {$\cdots$};

    \draw[<-] (2.6,-2.5) -- (2.75,-2.0);
    \draw[<-] (4.35,-2.5) -- (4.25,-2.0);

  
    \draw[<-] (2.4,-3.3) -- (2.55,-2.8);
    \node (p0_d) at (2.3,-2.9) {$\alpha_k$};

    \draw[<-] (4.6,-3.3) -- (4.45,-2.8);
    \node (p0_d) at (4.75,-2.9) {$\beta_k$};
    \filldraw[black] (4.6,-3.4) circle (2pt) node[anchor=west] {$r_{k+1}$};

  \filldraw[black] (2.4,-3.4) circle (2pt) node[anchor=west] {$s_{k+1}$};
\draw [->,decorate,decoration=snake] (2.4,-3.4) -- (2.4,-4.2);
  \filldraw[black] (2.4,-4.3) circle (2pt) node[anchor=north] {$g_{A}$};

   \draw[->,dashed,blue] (4.6,-3.4) to [bend left=30]  (2.45,-3.4);
   \draw[->,dashed,blue] (4.6,-3.4) to [bend right=30]  (2.8,-2.0);
   \draw[->,dashed,blue] (4.6,-3.4) to [bend right=0]  (2.7,-2.72);

  \filldraw[black] (7.0,0) circle (2pt) node[anchor=south] {s};

\node (p0_d) at (6.6,-0.4) {A};
  
  \filldraw[black] (6.5,-1.0) circle (2pt) node[anchor=east] {$s_1$};
  
  \draw[<-] (6.6,-0.9) -- (7.0,0.0);
  \draw[<-] (7.4,-0.9) -- (7.0,0.0);

  \filldraw[black] (7.5,-1.0) circle (2pt) node[anchor=west] {$r_1$};
\node (p0_d) at (7.4,-0.4) {B};

  \filldraw[black] (6.5,-4.3) circle (2pt) node[anchor=north] {$g_{A}$};

  \filldraw[black] (7.5,-4.3) circle (2pt) node[anchor=north] {$g_{B}$};
  
  \draw [->,decorate,decoration=snake] (6.5,-1.0) -- (6.5,-4.3);
  \draw [->,decorate,decoration=snake] (7.5,-1.0) -- (7.5,-4.3);
  

  \filldraw[black] (10.0,0) circle (2pt) node[anchor=south] {s};

\node (p0_d) at (9.6,-0.4) {A};
  
  \filldraw[black] (9.5,-1.0) circle (2pt) node[anchor=east] {$s_1$};
  
  \draw[<-] (9.6,-0.9) -- (10.0,0.0);
  \draw[<-] (10.4,-0.9) -- (10.0,0.0);

  \filldraw[black] (10.5,-1.0) circle (2pt) node[anchor=west] {$r_1$};
\node (p0_d) at (10.4,-0.4) {B};

  \filldraw[black] (9.5,-4.3) circle (2pt) node[anchor=north] {$g_{A}$};

  
  \draw [->,decorate,decoration=snake] (9.5,-1.0) -- (9.5,-4.3);
  \draw [->,decorate,decoration=snake] (10.5,-1.0) -- (11.0,-2.25);

\draw (11,-2.25) -- (12,-3.6) -- (10,-3.6) 
  -- cycle;

\node (a) at (0,-4.8) {(a)};
\node (b) at (3.5,-4.8) {(b)};
\node (c) at (7.0,-4.8) {(c)};
\node (d) at (10.0,-4.8) {(d)};

  
  
     

      
     
%

%
%
%
  \end{scope}
\end{tikzpicture}
  \caption{Possible plan behaviours after human-decision injection.}
  \label{fig:planbehaviour}
\end{figure*}
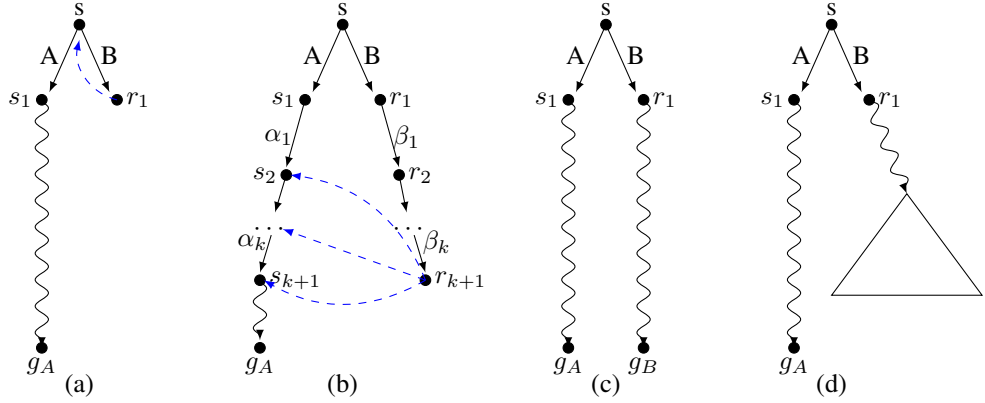

In order to evaluate different alternatives, it is necessary to infer by what metric the alternatives are to be compared (one plan might be longer but cheaper than a second --- depending on the relative values of time and money, either plan might be considered better). Furthermore, the user might want to change more than one action in the current plan, or iteratively revise the plan, or explore more than one alternative for a given action. This is represented by the diagram in Figure \ref{fig:planbehaviour} proposed in \cite{xaip}.  
Finally, as the set of possible alternatives the user might consider is potentially infinite, it is necessary to drive the user in the questions he/she can ask.

We considered all these issues in the development of {\sc XAI-Plan}, which is a methodology for providing explanations according to the view described above, and also gives the name to the framework implementing such methodology. 



\subsection{The {\sc XAI-Plan} Methodology}\label{sec:methodology}

{\sc XAI-Plan} is based on the idea that the user should be allowed to explore alternative plans by suggesting different actions in the plan. 
This paradigm provides an answer to questions like,
\begin{center}
\textit{Why does the plan contain action A rather than action B (that I would expect)?
}
\end{center}
\smallskip

The planning system is then used not only to generate the initial plan, but also to explore the alternative plans resulting from the user suggestions.

More formally, the {\sc XAI-Plan} methodology is presented in Algorithm~\ref{alg:pseudo}. This algorithm takes as input an initial set of plans, which at the beginning contains only a single plan. The user selects an action in the plan (line 2), and this corresponds to \textit{action A} in the question template above. As said before, the part of the question \textit{``rather than action B''} is open to a possibly infinite set of instances. To this end, {\sc XAI-Plan} restricts this set to the applicable actions (line 3). The list of applicable actions is generated in the following way: first the state in which action $a$ is applied in plan $\pi$ is obtained. Then, the set of all ground actions are filtered to only those whose preconditions are achieved in that state, minus the original action $a$. These actions are presented to the user, who can then select one of them (line 4).

\begin{algorithm}
\SetAlgoLined
\textbf{Input:} initial set of plans $\Pi$\\
 \Loop{}{
$\quad\rhd$ User chooses an existing plan\\
$1:\,\pi \leftarrow selectPlan(\Pi)$\\
$\quad\rhd$ User chooses an action within that plan\\
$2:\,a \leftarrow selectAction(\pi)$\\
$\quad\rhd$ Generate list of alternative actions\\
$3:\,applicable\_actions \leftarrow generateActions(\pi,a)$\\
$\quad\rhd$ User selects an alternative action\\
$4:\,a' \leftarrow selectAction(applicable\_actions)$\\
$\quad\rhd$ New plan is generated\\
$5:\,\pi' \leftarrow generatePlan(\pi,a,a')$\\
$6:\,\Pi \leftarrow \Pi\cup\pi'$
}
\caption{{\sc XAI-Plan}}
\label{alg:pseudo}
\end{algorithm}

Given a suggested action, a new plan is generated to answer the user's query (line 5). This can be done in one of four ways (described more in detail in the next section): planning from the initial state and forcing the user action to be performed, forcing the user action to be performed within  a time-window, planning from the state after applying the user action, or planning both plan segments before and after the user action separately.

Finally, the new plan is added to the list of plans $\Pi$, which can be compared, and selected for further modifications (line 6). This allows iterative exploration of alternative plans.

\subsection{Exploring Alternative Plans}
After the user selects an alternative action, one way to explore alternative plans is to inject the user action in the plan and then replan from there. Figure \ref{fig:planbehaviour} shows the  possible outcomes:
\begin{enumerate}[label=(\alph*)]
\item One possible behaviour is that the planner simply undoes the effect of the user action in order to return as quickly as possible to the original plan. While this might be the most efficient solution, it is undesirable from the standpoint of plan explanation, as it does not show clearly if an alternative exists, and the comparative quality of that alternative plan.
\item The reversal of the user action can be avoided by enforcing the planner does not revisit state $s$. The second behaviour illustrates that the new plan, through actions $B,\beta_1,\ldots,\beta_k$ does return to the original plan, by a more or less efficient route than actions $A,\alpha_1,\ldots,\alpha_k$. 
\item The third behaviour shows the case where a new plan is found by the planner, without returning to the original planned actions.
\item The fourth behaviour shows the case where no new plan is discovered, as the planner is unable to return to the original plan, and no alternative path to the goal exists.
\end{enumerate}

Note, however, that replanning after applying the user action is not the only option, as replanning from the initial state with additional constraints is also possible.
In the {\sc XAI-Plan} framework, there are four implementations of $generatePlan$ in Algorithm~\ref{alg:pseudo}. These are:
\begin{enumerate}
\item planning from the state after applying the user action,
\item planning from the initial state and forcing the user action to be performed,
\item planning from the initial state and forcing the user action to be performed within  a time-window,
\item or planning both plan segments before and after the user action separately.
\end{enumerate}

Planning from the state obtained after applying the user action is done by disallowing the undo action, and then replanning.
Planning from the initial state is achieved by updating the domain model to include a new predicate, {\small\verb|(applied-user-action)|}, which is included as an effect of a new operator {\small\verb|(user-action)|}. The new fact achieved by the user action is added as a goal of the problem, ensuring that the action is applied at least once in the plan.

When planning again from the initial state, there is a risk that the user action is performed differently from what was intended by the user. For example, the action might be appended to the end of a complete plan, simply to achieve the {\small\verb|applied-user-action|} effect. In a temporal plan, the user is able to specify a time-window in which they would like the action to be performed.
This is done by adding a new predicate to the domain that is a condition of the user action operator, and can be enabled and disabled by timed-initial-lieterals (TILs). For example, the fact {\small\verb|(applicable-user-action)|} is added as a new start condition of the operator {\small\verb|(user-action)|}. Then, two TILs are added to the problem,

{\small\verb|(at LB (applicable-user-action))|} and 

{\small\verb|(at UB (not (applicable-user-action)))|},
where LB is the lower bound on the time-window, and UB is the upper bound. This ensures that the action is applied in the time-window that interests the user.

The final strategy is to plan separately from the user action to the goal (the \emph{later plan}) and from the initial state to the user action (\emph{initial plan}). The final plan shown to the user is obtained by concatenating the initial plan, the user action, and the later plan. 
Planning the later plan is performed by planning from the state after applying the user action, using the original goals. Then, a new problem is generated with the same initial state as the original problem, and goal to achieve the weakest conditions of the later plan. The weakest conditions are those facts which are conditions for actions in the later plan, not already supported by effects. An example of each approach to plan generation is described in the next section.

\section{Examples of Exploring Alternative Plans}\label{sec:examples}

In this section we provide some examples of the four approaches to generating a new plan using the user suggested action, and some discussion on the strengths and drawbacks of each option.

\begin{figure}[thb]
\centering
\includegraphics[width=0.45\textwidth]{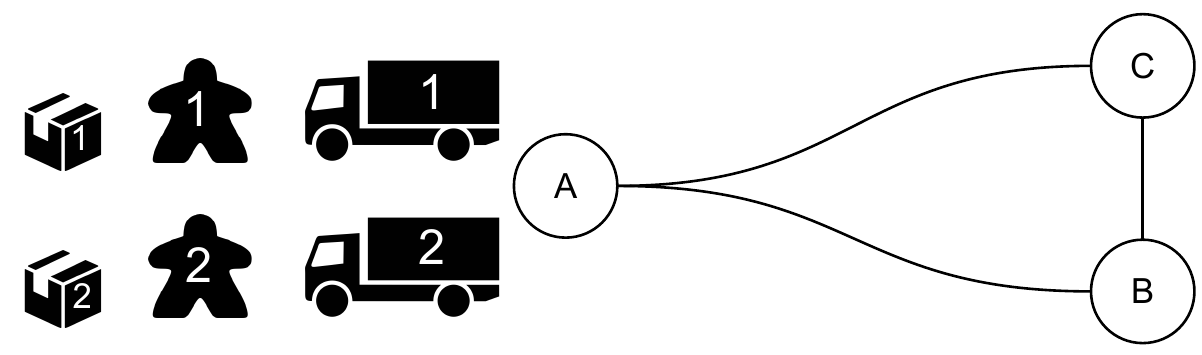}
\caption{Problem setup for the DriverLog domain. The drivers, trucks, and packages are at location A.}
\label{fig:driverlog1}
\end{figure}

Consider the problem shown in figure~\ref{fig:driverlog1} from the \textit{Driverlog} domain. In this problem two packages must be delivered to two separate locations. There are two drivers and two trucks available. Let us assume the plan found for this problem is shown in figure~\ref{fig:dl_plan1}. 

\begin{figure}[thb]
\small
\verb| 0.0:	(board-truck d2 t1 a) [10.0]|\\
\verb| 0.0:	(board-truck d1 t2 a) [10.0]|\\
\verb|10.0:	(load-truck p2 t1 a)  [10.0]|\\
\bverb|10.0:	(load-truck p1 t2 a) [10.0]|\\
\verb|20.0:	(drive-truck t2 a b d1) [30.0]|\\
\verb|20.0:	(drive-truck t1 a c d2) [30.0]|\\
\verb|52.0:	(unload-truck p1 t2 b)  [10.0]|\\
\verb|52.0:	(unload-truck p2 t1 c)  [10.0]|
\caption{Plan generated for the driverlog problem. The highlighted action is selected by the user to be changed.}
\label{fig:dl_plan1}
\end{figure}

We explore the case in which the user wishes to see plans that only use one truck to deliver the package, so that they might keep one truck in reserve.

\subsubsection{Example 1: Planning from the initial state}

The user selects the highlighted action in figure~\ref{fig:dl_plan1} and chooses the alternative action, {\small\verb|(load-truck p1 t1 a)|}, which loads the package into the first truck with the other package. A new plan must be generated from the initial state that achieves the goal and includes the action specified by the user.

This is done by updating the domain model to include a new predicate, {\small\verb|(applied-load-truck ?p ?t ?l)|}, which is included as an effect of a new operator {\small\verb|(user-action-load-truck)|}. The grounded fact {\small\verb|(applied-load-truck p1 t1 a)|}, achieved by the user selected action, is added as a goal of the problem. In general, the new operator included in the domain is an exact copy of the operator selected by the user, with the addition of the {\small\verb|user-action|} effect. The plan generated for this problem is shown in figure~\ref{fig:dl_plan2}.

\begin{figure}[thb]
\small
\verb| 0.0:	(board-truck d2 t1 a) [10.0]|\\
\verb| 0.0:	(board-truck d1 t2 a) [10.0]|\\
\verb|10.0:	(load-truck p2 t2 a)  [10.0]|\\
\bverb|10.0:	(load-truck p1 t1 a) [10.0]|\\
\verb|20.0:	(drive-truck t2 a c d1) [30.0]|\\
\verb|20.0:	(drive-truck t1 a b d2) [30.0]|\\
\verb|52.0:	(unload-truck p1 t1 b)  [10.0]|\\
\verb|52.0:	(unload-truck p2 t2 c)  [10.0]|
\caption{New plan generated from the initial state, forcing the highlighted user action.}
\label{fig:dl_plan2}
\end{figure}

When a time-window is specified, the user suggested {\small\verb|load-truck|} action must be applied within a time-window, as descibed in section~\ref{sec:methodology}. The resulting plan for this problem is the same as that shown in figure~\ref{fig:dl_plan2}.
As we can see, the planner swaps which package to load in which truck.

The drawback of planning from the initial state is clear from this example -- that the new plan still uses both trucks to deliver the packages. The question from the user, ``why not use only one truck to deliver both packages?'', is not given only by an action to be applied, but also the context in which that action should be applied. In this case, the action to load the package into truck {\small\verb|t1|} in the state in which the other package was already loaded into {\small\verb|t1|}.

\subsubsection{Example 2: Planning after the user action}

Planning from the state after applying the user action provides a plan that does show one truck delivering both packages, as shown in figure~\ref{fig:dl_plan3}

\begin{figure}[thb]
\small
\verb| 0.0:	(board-truck d2 t1 a) [10.0]|\\
\verb| 0.0:	(board-truck d1 t2 a) [10.0]|\\
\verb|10.0:	(load-truck p2 t1 a)  [10.0]|\\
\bverb|10.0:	(load-truck p1 t1 a) [10.0]|\\
\verb|20.0:	(drive-truck t1 a b d2) [30.0]|\\
\verb|52.0:	(unload-truck p1 t1 b) [10.0]|\\
\verb|62.0:	(drive-truck t1 b c d2) [30.0]|\\
\verb|94.0:	(unload-truck p2 t1 c) [10.0]|
\caption{Plan generated from the user action, using only one truck.}
\label{fig:dl_plan3}
\end{figure}

\begin{figure}[thb]
\centering
\includegraphics[width=0.45\textwidth]{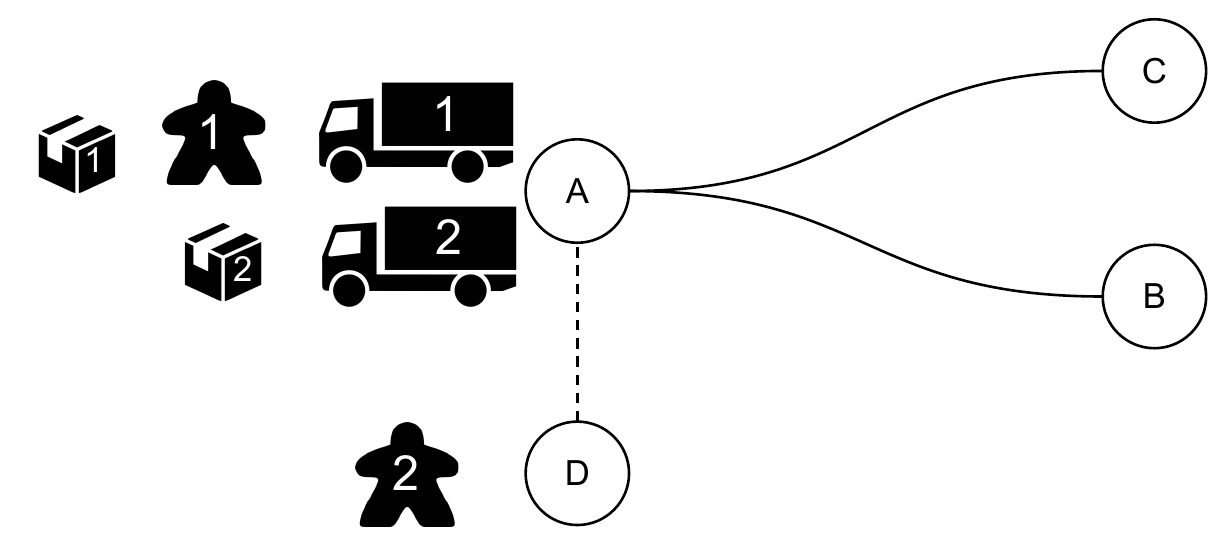}
\caption{Problem setup for the DriverLog domain. The drivers, trucks, and packages are at location A.}
\label{fig:driverlog2}
\end{figure}

Consider the problem as shown in figure~\ref{fig:driverlog2}. This problem resembles the first with the addition that one driver is separated from the trucks by a long path. A plan generated from the state after the user action is shown in figure~\ref{fig:dl_plan4}. The second driver, although no longer required, is still planned to walk along the long path to reach the trucks. This has the effect of dramatically reducing the plan's quality, and illustrates the drawback to planning from the user action -- the quality of the plan shown to the user might be far from what is realistic. Although the plan generation is answering the question the user asks by considering the suggested action and state in which it was suggested, the answering plan is not a good answer.

\begin{figure}[thb]
\small
\verb|  0.0:	(walk d2 d a) [60.0]|\\
\verb| 62.0:	(board-truck d2 t1 a) [10.0]|\\
\verb| 62.0:	(board-truck d1 t2 a) [10.0]|\\
\verb| 72.0:	(load-truck p2 t1 a)  [10.0]|\\
\bverb|72.0:	(load-truck p1 t1 a) [10.0]|\\
\verb| 82.0:	(drive-truck t1 a b d2) [30.0]|\\
\verb|114.0:	(unload-truck p1 t1 b) [10.0]|\\
\verb|144.0:	(drive-truck t1 b c d2) [30.0]|\\
\verb|176.0:	(unload-truck p2 t1 c) [10.0]|
\caption{Plan generated from the user action, using only one truck. The second driver  spends a long time walking.}
\label{fig:dl_plan4}
\end{figure}

\subsubsection{Example 3: Planning before and after the user action}

When planning the initial and later plans separately, the state in which the user suggested the action is not lost. The resultant plan, obtained by concatenating the initial plan, user action, and later plan is shown in figure~\ref{fig:dl_plan5}. This plan uses only one truck, as the later plan was generated from the state of the user's suggested action, and also does not waste time waiting for the second driver, as the initial plan was also regenerated.

\begin{figure}[thb]
\small
\verb| 0.0:	(board-truck d1 t1 a) [10.0]|\\
\verb|10.0:	(load-truck p2 t1 a)  [10.0]|\\
\bverb|10.0:	(load-truck p1 t1 a) [10.0]|\\
\verb|20.0:	(drive-truck t1 a b d1) [30.0]|\\
\verb|52.0:	(unload-truck p1 t1 b) [10.0]|\\
\verb|62.0:	(drive-truck t1 b c d1) [30.0]|\\
\verb|94.0:	(unload-truck p2 t1 c) [10.0]|
\caption{Plan generated from the user action, using only one truck, and without waiting for the second driver.}
\label{fig:dl_plan5}
\end{figure}

However, this approach also has the drawback that the context of the user action might be lost. For example, if the user suggested that the first load action be altered, then the action is suggested in the context of the later plan, and not the state in which the action is suggested. Identifying exactly what context matters with respect to an action suggestion is a challenging topic that we will explore in future work. 

\section{The \textsc{XAI-Plan} Framework}\label{sec:implementation}

\begin{figure*}[thb]
\begin{center}
\includegraphics[width=0.65\textwidth]{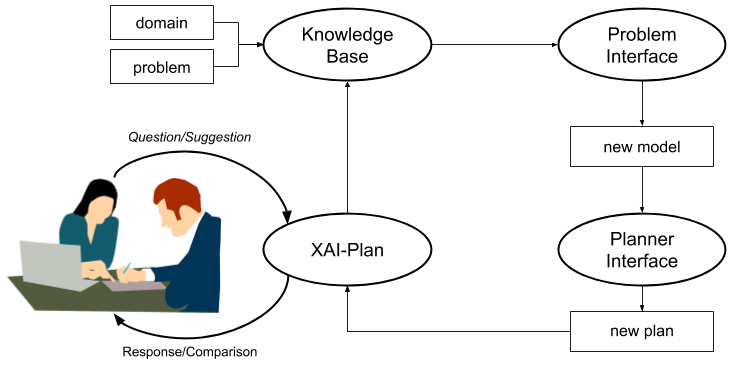}
\end{center}
\caption{{\sc XAI-Plan} architecture and its integration with ROSPlan.}
\label{fig:system}
\end{figure*}

{\sc XAI-Plan} has been implemented using the ROSPlan framework~\cite{rosplan15}, which provides an interface to AI planning systems and a method for storing and modifying PDDL models in the Robot Operating System (ROS).

Figure~\ref{fig:system} shows an overview of the {\sc XAI-Plan} framework. The XAI-Plan node implements the algorithms for generating explanatory plans, described in sections~\ref{sec:methodology} and \ref{sec:examples}, and communicates to the user through a user interface described below. The \emph{knowledge base}, \emph{problem interface}, and \emph{planner interface} are supplied by ROSPlan, which are used to store a PDDL model and provide an interface to the AI planner.

To provide a list of applicable actions to the user, the interface of the ROSPlan knowledge base is used to find the state at the point in the plan selected by the user, and then to retrieve the list of all grounded actions whose preconditions are achieved in that state.

Given an action suggested by a user, the domain is modified using the knowledge base interface, generating new predicates and operators that are required for the method of planning, as described in section~\ref{sec:examples}. The initial state and goal of each problem instance are also set using this interface. The \emph{problem interface} and \emph{planning interface} nodes are then called to generate new plans for these problem instances.

The new plans are concatenated, if required and as described in section~\ref{sec:examples}, and then shown to the user alongside a comparison with previously generated plans. The user controls these actions through a graphical user interface.

\begin{figure*}[!th]
\centering
\includegraphics[width=0.8\textwidth]{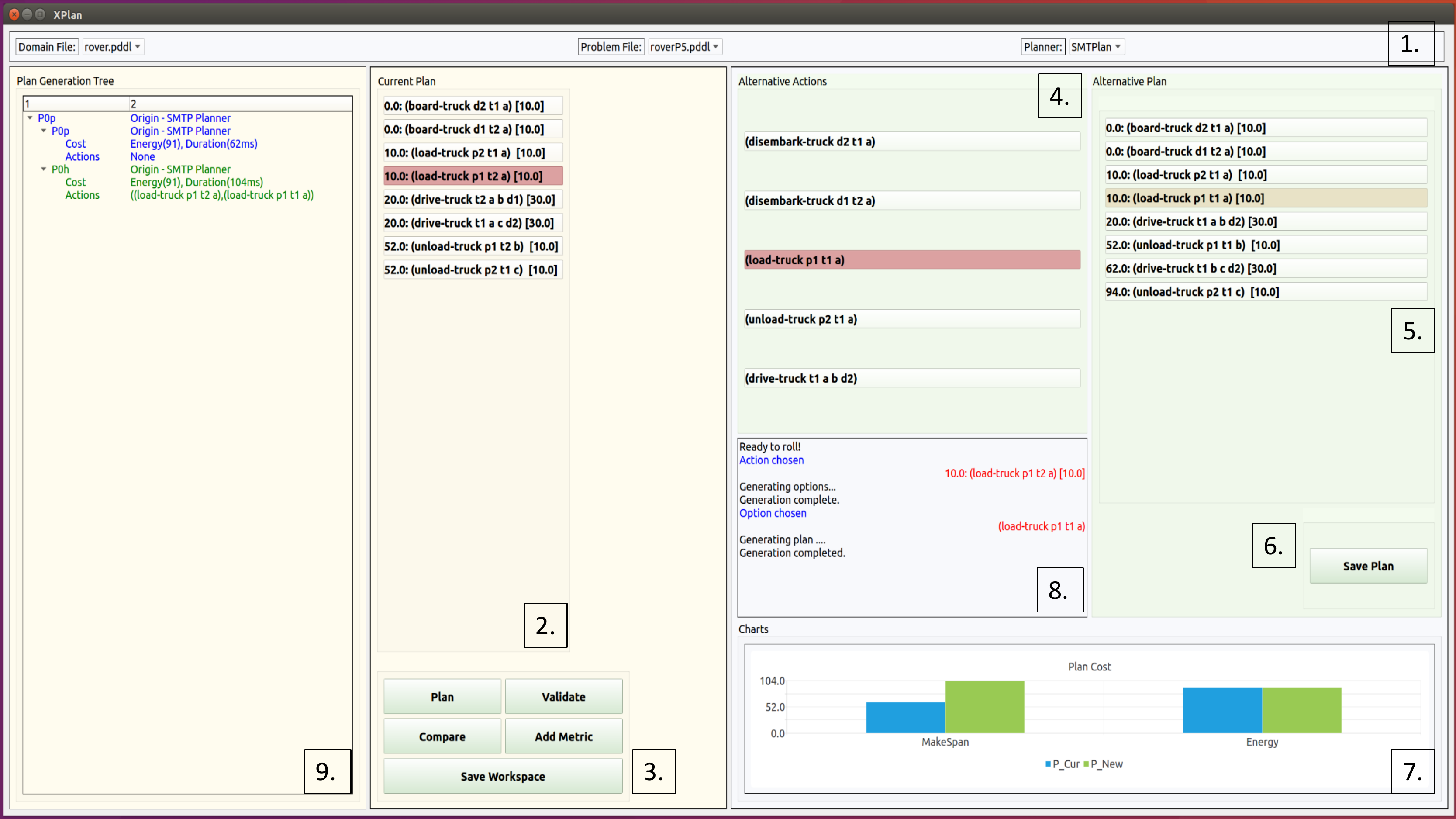}
\caption{XAI-Plan Interface detailed description. Actions highlighted in light red show the actions selected by the user for replacement. The action highlighted in light green shows the selected alternative action within a newly generated plan. }
\label{fig:gui}
\end{figure*}

\subsection{Interface Design}

The interface was designed following a user-centered approach; each component either supports a user initiated action or caters for a possible user need, such as performance comparison or plan generation tracking. A screenshot of the interface is provided in Figure~\ref{fig:gui}. The interface structure naturally unfolds the refinement cycles detailed in Algorithm~\ref{alg:pseudo}. 

A top bar ( \fbox{1} ) allows the user to select domain file, problem file, and planner. The \textit{Plan} button ( \fbox{3} ), when pressed, initiates the plan generation process using the selected planner.
The \textit{Current Plan} section ( \fbox{2} ) holds the current plan, which initially will be the one generated by the planner without user intervention.
Each action within the \textit{Current Plan} window can be individually selected by the user. After selection, {\sc XAI-Plan} will prompt the planner to suggest feasible alternative actions to the user. The list of alternative actions is displayed inside the \textit{Alternative Actions} section ( \fbox{4} ). The user is then allowed to either select any of the proposed alternative actions, or continue with the current plan and select a different action to compute a different set of alternative actions.
Selection of any action within the \textit{Alternative Actions} list will prompt the planner to generate an alternative plan which is displayed inside the \textit{Alternative Plan} section ( \fbox{5} ).
A textbox ( \fbox{8} ) provides real-time feedback to the user, the textbox also allows user annotation to be added to the communication stream.
Generated plans can be compared among each other via the \textit{Compare} button ( \fbox{3} ). The user is allowed to define metrics to evaluate the quality of a plan via the \textit{Add Metric} button ( \fbox{3} ). Quality of plans are displayed through a real-time chart in the \textit{Charts} section ( \fbox{7} ).
We have chosen a standard bar chart as the most appropriate representation for metric values. 
The \textit{Plans Generation Tree} section ( \fbox{9} ) encodes a "plan hierarchy".
The root of the hierarchy is the first plan generated by the planner. Each generated alternative plan saved by the user via the \textit{Save Plan} button ( \fbox{6} ) is automatically inserted into the hierarchy as a child of the current plan. 
Each node in the hierarchy contains: the plan ID, 
its cost, the action suggested by the user and the  action replaced. The tree allows an easy navigation between the plans the user is exploring.
Finally the \textit{Save Workspace} button ( \fbox{3} ) allows the user to save the current workspace.

\section{Conclusions}\label{sec:conclusion}

We have presented a methodology for providing explanations for the decisions made by the planner. The core idea is to allow the user to explore alternative actions within plans, generate a set of new plans, and then compare their costs. An explanation for a plan found by the planner is provided by showing that the plan is no worse than the alternative plan the user would expect. The methodology is implemented in the new \textsc{XAI-Plan} framework, which is integrated in ROSPlan, and provides a user-centered interface. Future work will explore a number of exciting issues such as temporal choices and context identification, and will feature a user study to assess the impact of explanations in trust and confidence.

\clearpage
\section*{Acknowledgements}
We thank David Aha, Maria Fox, and Derek Long for their very useful comments and suggestions.
\bibliographystyle{named}
\balance
\bibliography{bibliography}

\begin{thebibliography}{}

\bibitem[\protect\citeauthoryear{Bidot \bgroup \em et al.\egroup
  }{2010}]{biundospoken}
Julien Bidot, Susanne Biundo, Tobias Heinroth, Wolfgang Minker, Florian
  Nothdurft, and Bernd Schattenberg.
\newblock Verbal plan explanations for hybrid planning.
\newblock In {\em {MKWI}}, 2010.

\bibitem[\protect\citeauthoryear{Cashmore \bgroup \em et al.\egroup
  }{2015}]{rosplan15}
Michael Cashmore, Maria Fox, Derek Long, Daniele Magazzeni, Bram Ridder, Arnau
  Carrera, Narcis Palomeras, Natalia Hurtos, and Marc Carreras.
\newblock {ROSPlan}: Planning in the robot operating system.
\newblock In {\em {ICAPS}}, 2015.

\bibitem[\protect\citeauthoryear{Cashmore \bgroup \em et al.\egroup
  }{2018}]{auv}
Michael Cashmore, Maria Fox, Derek Long, Daniele Magazzeni, and Bram Ridder.
\newblock Opportunistic planning in autonomous underwater missions.
\newblock {\em {IEEE} Trans. Automation Science and Engineering},
  15(2):519--530, 2018.

\bibitem[\protect\citeauthoryear{Chakraborti \bgroup \em et al.\egroup
  }{2017}]{rao1}
T.~Chakraborti, S.~Sreedharan, Y.~Zhang, and S.~Kambhampati.
\newblock Plan explanations as model reconciliation: Moving beyond explanation
  as soliloquy.
\newblock In {\em {IJCAI}}, 2017.

\bibitem[\protect\citeauthoryear{Fox \bgroup \em et al.\egroup }{2017}]{xaip}
Maria Fox, Derek Long, and Daniele Magazzeni.
\newblock Explainable planning.
\newblock {\em IJCAI-17 Workshop on Explainable AI}, 2017.

\bibitem[\protect\citeauthoryear{Hayes and Shah}{2017}]{shah}
Bradley Hayes and Julie~A. Shah.
\newblock Improving robot controller transparency through autonomous policy
  explanation.
\newblock In {\em HRI}, 2017.

\bibitem[\protect\citeauthoryear{Katz \bgroup \em et al.\egroup }{2018}]{kat18}
Michael Katz, Shirin Sohrabi, Octavian Udrea, and Dominik Winterer.
\newblock A novel iterative approach to top-k planning.
\newblock In {\em ICAPS}, 2018.

\bibitem[\protect\citeauthoryear{Langley \bgroup \em et al.\egroup
  }{2017}]{langley17}
Pat Langley, Ben Meadows, Mohan Sridharan, and Dongkyu Choi.
\newblock Explainable agency for intelligent autonomous systems.
\newblock In {\em {AAAI}}, 2017.

\bibitem[\protect\citeauthoryear{Lipovetzky \bgroup \em et al.\egroup
  }{2014}]{pearceMining}
Nir Lipovetzky, Christina~N. Burt, Adrian~R. Pearce, and Peter~J. Stuckey.
\newblock Planning for mining operations with time and resource constraints.
\newblock In {\em {ICAPS}}, 2014.

\bibitem[\protect\citeauthoryear{Morris \bgroup \em et al.\egroup
  }{2015}]{morrisPlane}
Robert Morris, Mai~Lee Chang, Ronald Archer, Ernest Vincent~Cross II, Shelby
  Thompson, Jerry~L. Franke, Robert~Christopher Garrett, Waqar Malik, Kerry
  McGuire, and Garrett Hemann.
\newblock Self-driving aircraft towing vehicles: {A} preliminary report.
\newblock In {\em AAAI Workshop on AI for Transportation}, 2015.

\bibitem[\protect\citeauthoryear{Nguyen \bgroup \em et al.\egroup
  }{2012}]{ngu12}
Tuan Nguyen, Minh Do, Alfonso Gerevini, Ivan Serina, Biplav Srivastava, and
  Subbarao Kambhampati.
\newblock Generating diverse plans to handle unknown and partially known user
  preferences.
\newblock {\em Artificial Intelligence}, 2012.

\bibitem[\protect\citeauthoryear{Piacentini \bgroup \em et al.\egroup
  }{2016}]{UCP}
Chiara Piacentini, Daniele Magazzeni, Derek Long, Maria Fox, and Chris Dent.
\newblock Solving realistic unit commitment problems using temporal planning:
  Challenges and solutions.
\newblock In {\em {ICAPS}}, 2016.

\bibitem[\protect\citeauthoryear{Piotrowski \bgroup \em et al.\egroup
  }{2016}]{dino16}
Wiktor~Mateusz Piotrowski, Maria Fox, Derek Long, Daniele Magazzeni, and Fabio
  Mercorio.
\newblock Heuristic planning for {PDDL+} domains.
\newblock In {\em Proceedings of IJCAI}, 2016.

\bibitem[\protect\citeauthoryear{Rosenthal \bgroup \em et al.\egroup
  }{2016}]{veloso1}
Stephanie Rosenthal, Sai~P. Selvaraj, and Manuela~M. Veloso.
\newblock Verbalization: Narration of autonomous robot experience.
\newblock In {\em IJCAI}, 2016.

\bibitem[\protect\citeauthoryear{Seegebarth \bgroup \em et al.\egroup
  }{2012}]{biundo}
Bastian Seegebarth, Felix M{\"{u}}ller, Bernd Schattenberg, and Susanne Biundo.
\newblock Making hybrid plans more clear to human users - {A} formal approach
  for generating sound explanations.
\newblock In {\em {ICAPS}}, 2012.

\bibitem[\protect\citeauthoryear{Smith}{2012}]{smith12}
David Smith.
\newblock Planning as an iterative process.
\newblock In {\em AAAI}, 2012.

\bibitem[\protect\citeauthoryear{Sohrabi \bgroup \em et al.\egroup
  }{2011}]{Sohrabi}
Shirin Sohrabi, Jorge~A. Baier, and Sheila~A. McIlraith.
\newblock Preferred explanations: Theory and generation via planning.
\newblock In {\em {AAAI}}, 2011.

\bibitem[\protect\citeauthoryear{Sreedharan \bgroup \em et al.\egroup
  }{2018}]{rao3}
Sarath Sreedharan, Tathagata Chakraborti, and Subbarao Kambhampati.
\newblock Handling model uncertainty and multiplicity in explanations via model
  reconciliation.
\newblock In {\em ICAPS}, 2018.

\bibitem[\protect\citeauthoryear{Thi{\'{e}}baux \bgroup \em et al.\egroup
  }{2013}]{smartgridSylvie}
Sylvie Thi{\'{e}}baux, Carleton Coffrin, Hassan~L. Hijazi, and John~K. Slaney.
\newblock Planning with {MIP} for supply restoration in power distribution
  systems.
\newblock In {\em {IJCAI}}, 2013.

\bibitem[\protect\citeauthoryear{Vallati \bgroup \em et al.\egroup
  }{2016}]{vallatiUTC}
Mauro Vallati, Daniele Magazzeni, Bart~De Schutter, Luk{\'{a}}s Chrpa, and
  Thomas~Leo McCluskey.
\newblock Efficient macroscopic urban traffic models for reducing congestion:
  {A} {PDDL+} planning approach.
\newblock In {\em AAAI}, 2016.

\bibitem[\protect\citeauthoryear{Zhang \bgroup \em et al.\egroup }{2017}]{rao2}
Y.~Zhang, S.~Sreedharan, A.~Kulkarni, T.~Chakraborti, H.~Zhuo, and
  S.~Kambhampati.
\newblock Plan explicability and predictability for robot task planning.
\newblock In {\em {ICRA}}, 2017.

\end{thebibliography}

\end{document}